\documentclass[%
 aip,
 amsmath,amssymb,
reprint,%
]{revtex4-2}

\usepackage{graphicx}
\usepackage{dcolumn}
\usepackage{bm}

\usepackage[utf8]{inputenc}
\usepackage[T1]{fontenc}
\usepackage{mathptmx}
\usepackage{etoolbox}
\usepackage[mathscr]{euscript}

\usepackage[breaklinks]{hyperref}  
\hypersetup{
  colorlinks,
  citecolor=blue,
  linkcolor=blue,
  urlcolor=blue}
\usepackage[caption=false,position=top,margin=12pt,justification=raggedright,singlelinecheck=false,subrefformat=parens,labelformat=parens]{subfig}
\usepackage{subfig}
\makeatletter
\def\@email#1#2{%
 \endgroup
 \patchcmd{\titleblock@produce}
  {\frontmatter@RRAPformat}
  {\frontmatter@RRAPformat{\produce@RRAP{*#1\href{mailto:#2}{#2}}}\frontmatter@RRAPformat}
  {}{}
}%
\makeatother
\begin{document}


\title[Controlling Chaotic Maps using Next-Generation Reservoir Computing]{Controlling Chaotic Maps using Next-Generation Reservoir Computing}
\author{Robert M. Kent}
\altaffiliation{These authors contributed equally to this work}
\affiliation{ 
Department of Physics, The Ohio State University, 191 W. Woodruff Ave., Columbus, OH 43210, USA
}%

\email{kent.321@buckeyemail.osu.edu}
\author{Wendson A. S. Barbosa}%
\altaffiliation{These authors contributed equally to this work}
\affiliation{ 
Department of Physics, The Ohio State University, 191 W. Woodruff Ave., Columbus, OH 43210, USA
}%

\author{Daniel J. Gauthier}
\affiliation{ 
Department of Physics, The Ohio State University, 191 W. Woodruff Ave., Columbus, OH 43210, USA}
\affiliation{ResCon Technologies, LLC, 1275 Kinnear Rd., Suite 238, Columbus, OH 43212 USA
}%

\date{\today}

\begin{abstract}

In this work, we combine nonlinear system control techniques with next-generation reservoir computing, a best-in-class machine learning approach for predicting the behavior of dynamical systems. We demonstrate the performance of the controller in a series of control tasks for the chaotic H\'{e}non map, including controlling the system between unstable fixed points, stabilizing the system to higher order periodic orbits, and to an arbitrary desired state. We show that our controller succeeds in these tasks, requires only ten data points for training, can control the system to a desired trajectory in a single iteration, and is robust to noise and modeling error. 

\end{abstract}

\maketitle

\begin{quotation}
Developing effective control strategies for dynamical systems that display complex behaviors such as chaos poses significant challenges. Traditional control algorithms typically only work in a local linear neighborhood of a desired behavior or require an accurate system model, for example. Recent advances by the machine learning community provide new opportunities for controlling complex systems by using only measured data to identify an accurate model.  Recent advances include control algorithms that use artificial neural networks and model predictive control. While these are promising approaches, they can be computationally expensive and their complexity imposes limitations on hardware implementation.  In this study, we present an approach for controlling dynamical systems that harnesses the power of a novel machine learning method called next-generation reservoir computing. We show that our method can learn the underlying dynamics using only a few data points to learn the model and use it to control the system between unstable fixed points or arbitrary desired states even in the presence of noise and model error, thus displaying robustness.  
\end{quotation}

\section{\label{sec:Intro} Introduction}

\begin{figure*}[htb]
\centering
	\subfloat{%
	\includegraphics[width=\linewidth]{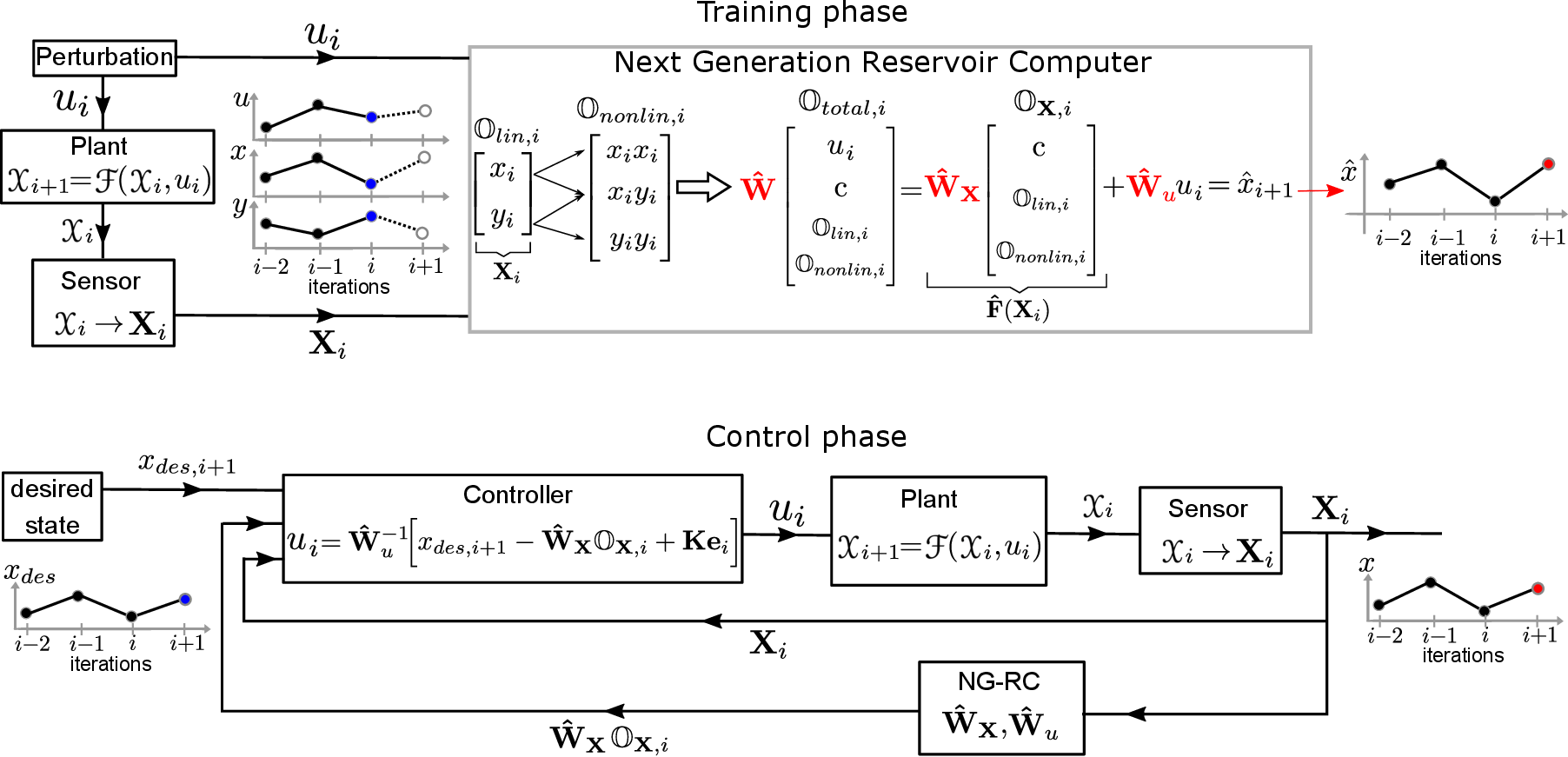}%
	}
\caption{Learning and controlling a dynamical system (plant) using next-generation reservoir computing. (top) The NG-RC is trained to predict $x_{i+1}$, the next step of one of the plant's observable variable. (bottom) The controller is designed based on the NG-RC model estimation for the evolution of the observable variable to be controlled. Then, it applies a control signal $u_i$ to the plant, which is driven towards a desired state $x_{des,i+1}$.}
\label{fig:Storyboard}
\end{figure*}

Controlling dynamical systems is a problem of wide importance as it encompasses various domains including robotics, aerospace, industrial processes, and biological systems.  The goal is to design an algorithm that forces the system, known as a \textit{plant}, to follow a prescribed trajectory in phase space that is robust against noise and other external influences.  Controlling a system requires, at a minimum, the existence of an accessible system parameter that modifies the dynamics.  The accessible parameter is adjusted based on a control algorithm, of which there are two general classes: open-loop algorithms where the adjustment is made independent of the state of the system; and closed-loop controllers that adjust the control signal based on real-time measurement of the state of the system.  We focus here on closed-loop controllers, an example of which is illustrated in the bottom panel of Fig.~\ref{fig:Storyboard}.

A particularly challenging control problem is when the system's dynamics have nonlinear interactions that give rise to complex behaviors such as chaos. Typically, controlling nonlinear systems requires an accurate mathematical model of the system if control is desired over the entire phase space and range of operating parameters.\cite{Slotine1991} 

The study of controlling chaotic dynamical systems was transformed in the 1990s by the realization that the unstable periodic orbits and other unstable sets that are the skeleton of the chaotic attractor can be used as the basis of a control algorithm.\cite{Ott_Grebogi_Yorke1990}  Here, only the local linear dynamics of the system in the vicinity of a periodic orbit are required, and hence, linear control methods\cite{Ogata} can be used.  The system naturally visits the periodic orbits, and control is only applied when there is a close approach; hence, only small adjustments to the accessible system parameter are required.  Many variations on this theme have been developed and applied to a wide variety of engineered and natural systems.\cite{Weeks_1997,Gauthier2003}

There is a resurgence in research on nonlinear control that can stabilize arbitrary states by bringing together traditional nonlinear control methods and modern machine learning (ML) tools that are used to learn a data-driven model of the system.  Examples include model predictive control,\cite{Camacho2013,Wu2020} control using artificial neural networks,\cite{Poznyak_1999,Zong2009,Gokce2013} evolutionary algorithms, \cite{Matousek2019,Matousek2021,Senkerik2011,Zelinka2009,Matousek2014} and recurrent neural networks,\cite{Chow_1998} among others. However, these methods require large sets of data for creating the model (called \textit{training data}), and they require high-complexity model evaluation, which limits their use for controlling high-speed systems or deployment on low-power microcontrollers (known as edge computing devices).  

A promising ML algorithm for controlling dynamical systems that may offer a solution to these bottlenecks is reservoir computing.\cite{Jaeger_Haas2004,Maass2002} It is a best-in-class method for identifying data-driven models of dynamical systems and has already been applied successfully to a few problems.\cite{Waegeman2012,Waegeman2013,Canaday2021,Haluszczynski2021} Combining reservoir computing with time delays can reduce the model size, \cite{KurthsDelayRC2023} but there is a need to further reduce the algorithm complexity. Next-generation reservoir computing\cite{NGRC} is mathematically equivalent to traditional reservoir computing but requires less training data, is less computationally expensive, and has fewer hyperparameters to optimize. The next-generation reservoir computing paradigm presents state-of-the-art prediction for low- \cite{NGRC} and high-dimensional\cite{NGRC-L96} chaotic dynamical systems. A controller based on next-generation reservoir computing has been used to restore the original behavior of a dynamical system affected by external factors or order parameter changes, and was found to require substantially less training data than its classical counterpart.\cite{haluszczynski_2023} Our controller design is novel in that it is not restricted to chaotic, intermittent, or periodic target trajectories present in the original dynamics of the system. Instead, we control the system to trajectories not belonging to the underlying attractor, such as transitioning between two unstable fixed points, and arbitrary points that lie outside the bounding region of the attractor.

The primary goal of this paper is to demonstrate how reservoir computers, including the next-generation reservoir computer (NG-RC), can be placed in the framework of control algorithms developed by Sarangapani.\cite{Sarangapani2006}  This general class of control algorithms can be applied to discrete dynamical systems using what is called a self-tuning regulator and can also be applied to continuous-time systems that are sampled discretely. Previous work involving neural networks has demonstrated that control is possible for continuous-time systems when some parameters\cite{Poznyak_1999} or system variables are not measured,\cite{Kim_1999} but the sampling time must be very short and the weights must be updated every step, which is computationally expensive. In contrast, Sarangapani considers control of discrete maps using a trained artificial neural network that is a linear sum of nonlinear functionals of the input data and additive control perturbations, known as the discrete Brunovsky canonical form of the estimated system.  Note that the underlying map does not need to be of this form, only the learned model.

Sarangapani's framework is a suitable starting point because the output of a reservoir computer is a linear sum of nonlinear functionals of the input data.  Importantly, Sarangapani proves, using Lyapunov stability analysis, that any discrete system in a Brunovsky form can be controlled robustly as long as the ML model error and system noise are bounded.  To illustrate this approach with a pedagogical example, we show how an NG-RC can be used to control the H\'{e}non map\cite{Henon1976} by using it to stabilize unstable periodic orbits and to also guide the system to arbitrary dynamical states.  We show that control is robust in the presence of noise and modeling error, the NG-RC can be trained to faithfully reproduce the behavior of the H\'{e}non map using only a few training data points, and the learned model can be evaluated by performing a small number of computations.  

The control scheme concept is illustrated in Fig. \ref{fig:Storyboard}, which has two distinct phases. In the first phase, the NG-RC is trained to predict the next step of the system's observable variable to be controlled, taken to be a single variable $x$ in this illustration. The NG-RC predicts $x_{i+1}$ having as input the system's observable variables, taken as  ${\bf{X}}_i=\{x_i,y_i\}$ for this illustration, which are measured through a sensor connected to the plant (dynamical system).  An external perturbation $u_i$ at the current time step $t_i$, also taken as a scalar in this illustration, is applied directly to the plant. The input data (blue circles) to the NG-RC are used to create a feature vector $\mathbb{O}_{total}$ that includes $u_i$, a constant $c$, and linear ($\mathbb{O}_{lin,i}$) and nonlinear ($\mathbb{O}_{nonlin,i}$) terms derived from ${\bf{X}}_i$. 

The NG-RC is trained by using a random perturbation sequence $u_i$ (length $M_{train}$), which teaches the algorithm the underlying dynamics of the discrete-time system and its response to perturbations.  The NG-RC is trained so that its output is equal to the measured accessible variable, in a least-square sense, given by $x$.  This optimization procedure identifies ${\bf{\hat W}}$ that minimizes the difference between the true  $x_{i+1}$ and the predicted variable $\hat x_{i+1}$ (see the Appendix \ref{app:OptimizingAlpha}).

In the control phase, a controller is used to generate the signal $u_i$ that is applied to the plant. The control signal is designed to drive the plant to the next state $x_{i+1}$ toward the desired state $x_{des,i+1}$ one step in the future. The controller includes information of $x_{des,i+1}$, ${\bf{X}}_i$, and NG-RC prediction of the uncontrolled system ${\bf{\hat W}}_{\bf{X}}{\mathbb{O}}_{{\bf{X}},i}$.

The rest of the paper is organized as follows. In Sec. \ref{sec:Control}, we describe the general formulation for a dynamical system whose evolution is described by a discrete map, followed by the ML-based control law formalism used in this work. In Sec. \ref{sec:NGRC}, we introduce the theoretical background of the NG-RC and describe how well it suits the proposed control algorithm. In Sec. \ref{sec:HenonMap}, we describe the dynamical system used as a plant for our proposed control method, which details on implementation is shown in Sec. \ref{ControllingHenonMap}. Section \ref{sec:Results} is dedicated to the results. Here, we use our method to control the H\'{e}non map to three different desired trajectories. We show that our algorithm can correctly learn the system in the presence of noise, thus showing to be a robust controller. Last, we close the paper with discussions, conclusions, and future directions in Sec. \ref{sec:Discussions}.  

\section{\label{sec:Control} Controlling Discrete Dynamical Systems Using Machine Learning}  

In this section, we describe Sarangapani's \cite{Sarangapani2006} nonlinear control framework for discrete-time systems. First, we describe the general formulation for a multi-input multi-output (MIMO) dynamical system subject to disturbances, whose evolution is described by a discrete map. Then, we formulate the control algorithm. 

\subsection{Dynamical system evolution in the presence of feedback perturbations} 

Controlling a dynamical system relies on designing a control perturbation ${\bf{u}}$ applied to an accessible system parameter that drives the system to a desired behavior. The system's evolution is described in general by the map 
 \begin{eqnarray}
 {\bf{\mathscr{X}}}_{i+1}&=&{{\bf{\mathscr{F}}}}\left({\bf{\mathscr{X}}}_{i},{\bf{u}}_i\right), 
\label{eq:Xreal} 
\end{eqnarray}
where the system variables ${\bf{\mathscr{X}}}_i\in \mathbb{R}^{d_s}$ represent the system state at time $t_i$; $d_s$ is the system's phase-space dimension; and ${\bf{\mathscr{F}}}(\cdot)$ is a nonlinear function, generally unknown, that maps (${\bf{\mathscr{X}}}_i,{\bf{u}}_i$) to ${\bf{\mathscr{X}}}_{i+1}$. In real-world systems, it can be difficult to measure all components of ${\bf{\mathscr{X}}}_i$ because some may not be accessible or it is impractical to measure all components if the system is high dimensional. Thus, the observed dynamics are a projection of the real dynamics in the state-space formed by the set of accessible observations ${\bf{X}}$, whose evolution is given by

 \begin{eqnarray}
{\bf{X}}~=~\bf{\mathcal{G}}({\bf{\mathscr{X}}}),
\label{eq:X}
\end{eqnarray}
where ${\bf{X}}_{i} \in \mathbb{R}^{d'}$ with $d' \le d_s$ and $\bf{\mathcal{G}}$ is a possibly nonlinear function. In this work, we assume that all variables are observed, but this framework allows for the estimation of unmeasured variables using observers,\cite{Sarangapani2006} which is not unique to this approach.\cite{Kim_1999} Alternatively, $\bf{\mathcal{G}}$ can be approximated using techniques, such as time-delay embedding \cite{takens_1981} used to characterize dynamical systems from limited measurements.

A second consideration is that the control perturbation ${\bf{u}}$ might be applied to a subset ${\bf{Y}} \in \mathbb{R}^{d}$ of the observed variables ${\bf{X}}$ so that ${\bf{u}}$ is a ${d}$-dimensional vector. Here, ${\bf{Y}}$ represents the variables to be controlled whose dynamics are described by 
\begin{eqnarray}
{{\bf{Y}}_{i+1}} &=& {\bf{\mathcal{F}}}({\bf{X}}_i,{\bf{u}}_i),
\label{eq:Y}
\end{eqnarray}
where ${\bf{\mathcal{F}}}$ is a nonlinear function, generally unknown, that maps $({\bf{X}}_i,{\bf{u}}_i)$ to ${{\bf{Y}}_{i+1}}$. 

The control problem addressed in this paper relies on finding the control signal ${\bf{u}}$ that minimizes ${\bf{Y}}_{i+1}-{\bf{Y}}_{des,i+1}$, where ${\bf{Y}}_{des,i+1}$ is the desired state one step in the future. 

\subsection{\label{sec:ControlLaw}Control Law Design}

In this work, we assume that the perturbations enter linearly, which occurs for a wide class of dynamical systems, so that
 \begin{eqnarray}
 {\bf{\mathcal{F}}}({\bf{X}}_i,{\bf{u}}_i)&=&{\bf{F}}\left({\bf{X}}_{i}\right) + {\bf{W}}_u{\bf{u}}_i.
\label{eq:F_as_F_and_Wu}
\end{eqnarray}
Here, ${\bf{W}}_u \in \mathbb{R}^{d\times d}$ is a matrix whose components represent the control effectiveness and ${\bf{F}}: \mathbb{R}^{d'} \rightarrow \mathbb{R}^{d}$ is the map of the uncontrolled $d$-dimensional system. Thus, the dynamical system represented in Eq. \ref{eq:Y} can be described by
 \begin{eqnarray}
 {\bf{Y}}_{i+1}&=&{\bf{F}}\left({\bf{X}}_{i}\right) + {\bf{W}}_u{\bf{u}}_i + {\bf{d}}_i,
\label{eq:controllinearu}
\end{eqnarray}
where {\bf{d}} is a time-dependent disturbance that arises, for example, from noise. The goal of a ML algorithm is to learn a model $\hat{\bf{F}}$ of the underlying map $\mathbf{F}$ and $\hat{\bf{W}}_u$ for the control effectiveness.  It is possible to generalize Eqs.~\ref{eq:F_as_F_and_Wu} and \ref{eq:controllinearu} to have a nonlinear functional of $\mathbf{X}_i$ multiplying $\mathbf{u}$,\cite{Sarangapani2006} but we restrict our study to the simpler case to illustrate the approach with as little complexity as possible.


To arrive at a control algorithm, consider the controller output tracking error at time $t_{i+1}$ defined as the difference between the actual and desired outputs 
 \begin{eqnarray}
 {\bf{e}}_{i+1} \equiv {\bf{Y}}_{i+1} - {\bf{Y}}_{des,i+1}.
\label{eq:error}
\end{eqnarray}
Using this definition in Eq. \ref{eq:controllinearu}, the dynamics of the output tracking error are given by  
  \begin{eqnarray}
 {\bf{e}}_{i+1} &=&{\bf{F}}({\bf{X}}_{i}) + {\bf{W}}_u{\bf{u}}_i - {\bf{Y}}_{des,i+1} + {\bf{d}}_i.
\label{eq:errorv2}
\end{eqnarray}
Sarangapani\cite{Sarangapani2006} demonstrates that a robust and provable stable nonlinear controller can be realized by solving for $\mathbf{u}_i$ in Eq.~\ref{eq:errorv2}, setting $\mathbf{d}_i = 0$, and replacing $\mathbf{e}_{i+1} \rightarrow \mathbf{Ke}_i$, $\mathbf{F} \rightarrow \hat{\mathbf{F}}$, and $\mathbf{W}_{\mathbf{u}} \rightarrow \hat{\mathbf{W}}_{\mathbf{u}}$.  The control perturbations are then given by 
 \begin{eqnarray}
{\bf{u}}_i&=&\hat{{\bf{W}}}_u^{-1}\left[{\bf{Y}}_{des,i+1} - \hat{\bf{ F}}({\bf{X}}_{i}) + {\bf{K}} {\bf{e}}_{i} \right], 
\label{eq:controlsignal}
\end{eqnarray}
where ${\bf{K}}$ is the closed-loop gain matrix. Using this form of ${\bf{u}}$, the tracking error dynamics becomes 
 \begin{eqnarray}
 {\bf{e}}_{i+1}&=&{\bf{F}}({\bf{X}}_{i})  + {\bf{W}}_u {\hat{{\bf{W}}}_u}^{-1}\left[{\bf{Y}}_{des,i+1} - \hat{\bf{ F}}({\bf{X}}_{i}) + {\bf{K}} {\bf{e}}_i \right] \nonumber \\
 &~~& -{\bf{Y}}_{des,i+1} + {\bf{d}}_ i. 
\label{eq:error2}
\end{eqnarray}

Equation~\ref{eq:error2} is informative: the control perturbations attempt to cancel the nonlinear function $\mathbf{F}$ via the term $\hat{\mathbf{F}}$, referred to as feedback linearization, and simultaneously apply linear proportional feedback via the term $\mathbf{Ke}_i$. In the ideal situation where the map and control effectiveness are learned without error, i.e., $\hat{\bf{F}}({\bf{X}}_{i}) = {\bf{F}}({\bf{X}}_{i})$ and ${\hat{{\bf{W}}}_u} = {\bf{W}}_u $, and there is no noise, the tracking error takes the simple form
\begin{equation}
{\bf{e}}_{i+1} = {\bf{K}}{\bf{e}}_i.  
\label{eq:error_perfect}
\end{equation}
The tracking error approaches zero when the Floquet multipliers of map (\ref{eq:error_perfect}) are within the unit circle.  Floquet multipliers close to the unit circle give a slower decay of the tracking error, whereas values close to zero give rise to faster decay.  

In Sec. \ref{sec:NGRC}, we show that the NG-RC meets the criteria for such a machine-learning-based control algorithm and we describe how to construct the NG-RC, train it for control, and implement it in this control framework.

\section{\label{sec:NGRC} Next-Generation Reservoir Computing}

In this section, we show how to estimate ${\bf{F}}$ and ${\bf{W}}_u$ using a next-generation reservoir computer.\cite{NGRC} This is the training phase shown in the top panel of Fig.~\ref{fig:Storyboard}. The NG-RC is trained with supervised learning, where input data and target output are known for a number of iterations $M_{train}$ and the perturbations are generated from a normally distributed random sequence with zero mean and standard deviation $\sigma_{\textbf{u}}$.

Similar to other ML approaches, the core tasks are to design the feature vector  $\mathbb{O}_{total}$, 
whose linear transformation expresses the ML output as
 \begin{eqnarray}
 \hat{{\bf{Y}}}_{i+1} &=& \hat{{\bf{ W}}}\mathbb{O}_{total,i},
\label{eq:MLoutput}
\end{eqnarray}
and optimizing the output matrix of coefficients $\hat{{\bf{W}}}$ for minimizing the difference  $||\hat{{\bf{Y}}}_{i+1} - {\bf{Y}}_{i+1}||$ in a least-square sense (see the Appendix \ref{app:OptimizingAlpha}) between the predicted and true variables, thus maximizing the model performance. In traditional ML methods, such as RCs, $\mathbb{O}_{total}$ is a functional representing the nonlinear transformation of the input data performed by an artificial neural network. In the NG-RC formalism, $\mathbb{O}_{total}$ includes the control signal and a functional composed of linear and nonlinear parts that act directly on the input data. To respect the assumption that the control perturbation enters linearly, we exclude ${\bf{u}}_i$ from the nonlinear parts as follows:
 \begin{eqnarray}
 \mathbb{O}_{total,i}&=& {\bf{u}}_i \oplus c \oplus \mathbb{O}_{lin,i} \oplus  \mathbb{O}_{nonlin,i}.
\label{eq:Feats}
\end{eqnarray}
Here, $\oplus$ is the concatenation operator and $c$ is a constant. The linear part is a vector with dimension $d_{lin}=d'$ and is given by 
 \begin{eqnarray}
 \mathbb{O}_{lin,i}&=  {\bf{X}}_i,
\label{eq:lin feat observable variables}
\end{eqnarray}
and, hence, includes the observable system variables in the current time step. 

While there is great flexibility in selecting the functions that compose the nonlinear part $\mathbb{O}_{nonlin}$, we use the unique order-$p$ monomials formed by the components of $\mathbb{O}_{lin,i}$. This choice is motivated by the realization that an arbitrary dynamical system can be expressed by a Volterra series.\cite{Boyd1985,Gonon2022} Because the system that we control in this work has quadratic polynomial terms, we expect that the series can be truncated at quadratic order, but system identification techniques\cite{Wei_and_Billings_2004} can be used when little information is known about the system, which is discussed in Sec. \ref{ControllingHenonMap}. Similar aggressive truncation of a polynomial functional representation has been used previously to obtain state-of-the-art performance for dynamical system prediction.\cite{NGRC,NGRC-L96} Thus, the final dimension of $\mathbb{O}_{nonlin}$ is given by $d_{nonlin}=d_{lin}(d_{lin}+1)/2$, and the dimension of $\mathbb{O}_{total}$ is $d_{tot} = 1 + d + d_{lin} + d_{nonlin}$.

To meet the criteria imposed by Sarangapani's control framework presented in Sec. \ref{sec:Control} that the learned model needs to be in the Brunovsky canonical form, i.e., a linear sum of nonlinear functionals of the input data and additive control perturbations, we rewrite the NG-RC output as
 \begin{eqnarray}
 \hat{{\bf{Y}}}_{i+1} &=& {\bf{\hat W}}_{\bf{X}}\mathbb{O}_{{\bf{X}},i} + {\bf{ \hat W}}_u{\bf{u}}_i,
\label{eq:MLoutputWxWu}
\end{eqnarray}
where ${\bf{ \hat W}}_u$ is the learned weight matrix for the perturbation ${\bf{u}}$ and ${\bf{\hat W}}_{\bf{X}}$ is the learned weight matrix for the set of features originated from ${\bf{X}}_i$ plus the constant, which is given by $\mathbb{O}_{{\bf{X}},i} = c \oplus \mathbb{O}_{lin,i} \oplus  \mathbb{O}_{nonlin,i}$. Finally, Eq. \ref{eq:MLoutputWxWu} can be written in the format of Eq. \ref{eq:controllinearu} as
 \begin{eqnarray}
 \hat{{\bf{Y}}}_{i+1} &=& {\bf{\hat F}}({\bf{X}}_i) + {\bf{ \hat W}}_u{\bf{u}}_i + {\bf{\epsilon_i}},
\label{eq:MLoutputFxWu}
\end{eqnarray}
where ${\bf{\hat F}}({\bf{X}}_i) = {\bf{\hat W}}_{\bf{X}}\mathbb{O}_{{\bf{X}},i}$ is the NG-RC next step prediction of the unperturbed plant and ${\bf{\epsilon_i}}$ is the machine learning modeling error, also known as parameter estimation error.  

In this work, we focus on controlling a low-dimensional system, but higher-dimensional systems may pose additional challenges due to situations where the number of system variables that can be measured and accessible system parameters that can be adjusted are limited. Neural-network based observers have been proposed to infer the dynamics of unmeasured variables in controllers for large-scale systems,\cite{Tong_2011} but have only been applied to a low-dimensional problem. It has been shown that an NG-RC can learn accurate models of high-dimensional systems that exhibit complex dynamics even when only a subset of variables are measured.\cite{NGRC-L96} However, it remains an open research question on how to scale our controller to these high-dimensional problems with limited observations. 

\section{\label{sec:HenonMap}The H\'{e}non Map}

\begin{figure}[!ht]
\centering
	\subfloat{%
	\includegraphics[scale=0.42]{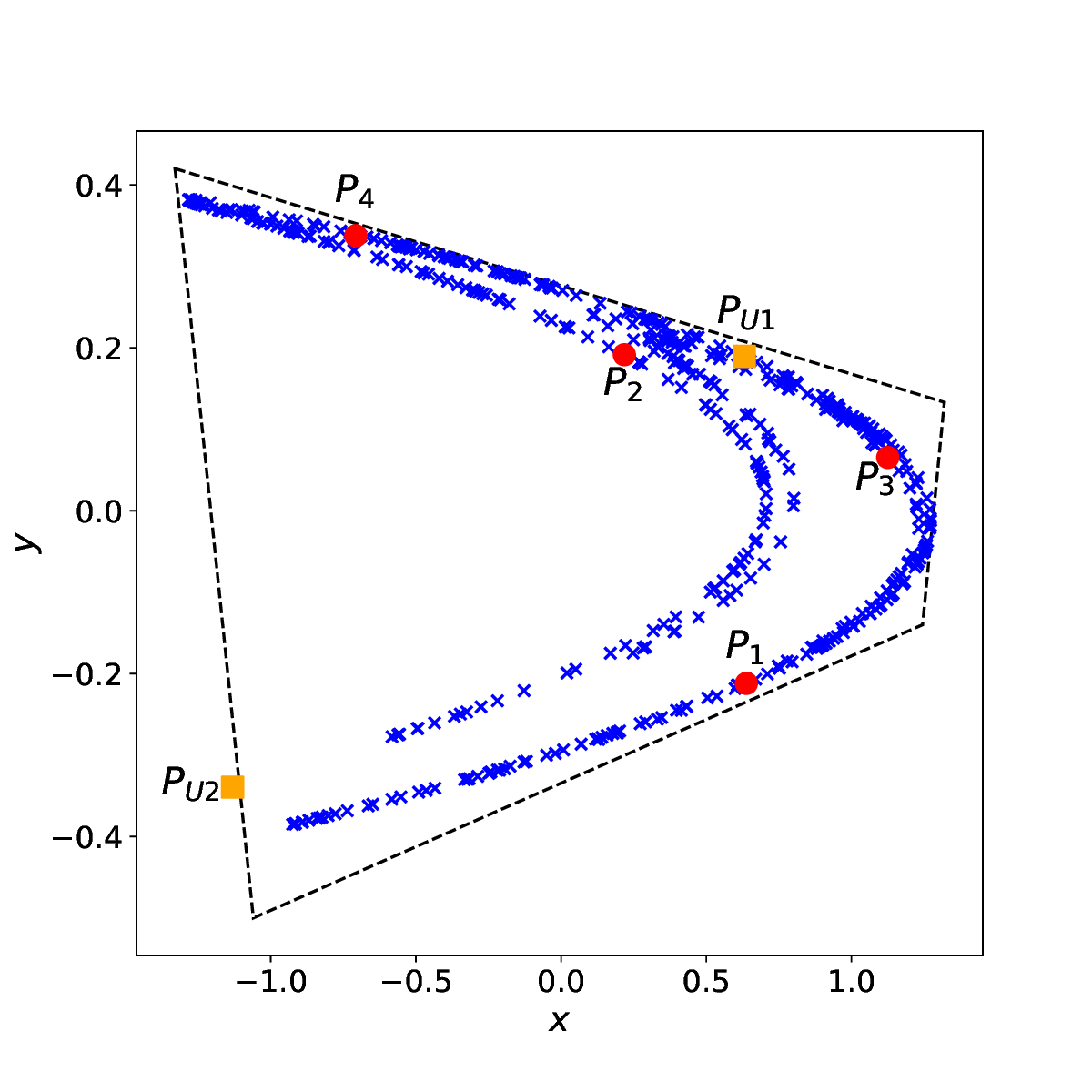}%
	}

\caption{Phase portrait of the H\'{e}non attractor for $a=1.4$, $b=0.3$ iterated for 500 points (blue xs). The orange square represents the unstable fixed points $P_{U1}$ and $P_{U2}$ with coordinate values $(x_{U1},y_{U1}) = (+0.63135,+0.18941)$ and $(x_{U2},y_{U2}) = (-1.13135,-0.33941)$, respectively. The red circles represent the points in the four-period orbit $P_1,P_2,P_3,P_4$ given by $(x_1,y_1)=(0.638194,-0.21203)$, $(x_2,y_2)=(0.217762,0.191458)$, $(x_3,y_3)=(1.12507,0.0653285)$, and $(x_4,y_4) = (-0.706767,0.337521)$, respectively. The dashed black line represents the trapping region: all points within the region remain within the region, and all points outside diverge to infinity.\cite{Henon1976}}
\label{fig:attractor}
\end{figure}

The H\'{e}non map \cite{Henon1976} is a simplified model of a continuous-time weather model that displays chaos\cite{Lorenz1963} while still displaying many of the same essential properties. It has been widely used to study chaotic behavior, such as chaos synchronization,\cite{Tutueva2022} anti-control,\cite{Han2021} image encryption,\cite{Soleymani2014,Khan2015} random number generation,\cite{Vajargah2015} and secure communication.\cite{Belkhouja2018} The map is defined by
\begin{align}
    x_{i+1} &= 1 - a x_{i}^{2} + y_{i} + g u_{i} + d_{x,i} \label{eq:henon map xip1}\\
    y_{i+1} &= b x_{i} + d_{y,i} \label{eq:henon map yip1},
\end{align}
where H\'{e}non chose parameters $a=1.4$, and $b=0.3$ for which the map exhibits chaotic behavior, where $g$ is the gain on the control perturbation $u_i$, and $d_{x,i}$ and $d_{y,i}$ are additive noise terms. For now, we take $u_{i} = d_{x,i} = d_{y,i} = 0$ to attain the classic equation. Throughout the rest of the paper, we use the original values of $a$ and $b$. A phase-space trajectory of the H\'{e}non map can be seen in Fig. \ref{fig:attractor}.

The H\'{e}non map has two unstable fixed points, $P_{U1}$ and $P_{U2}$. $P_{U1}$ is inside the trapping region that bounds the attractor, while $P_{U2}$ is located outside the trapping region. Any points inside the trapping region remain inside, and any point outside will diverge to infinity (see Fig. \ref{fig:attractor}). 

The unstable fixed points can be found by solving Eq.~\ref{eq:henon map xip1} and Eq.~\ref{eq:henon map yip1} for $x_{i+1}=x_n$ and $y_{i+1}=y_n$, respectively. When the system is in one of these fixed points, it is said to be in a one-period orbit because it repeats with every iteration of the map. Higher orbits, such as two-period and four-period, can be computed using a similar approach. The points in the four-period orbit are shown in Fig. \ref{fig:attractor}.

\section{\label{ControllingHenonMap} Controlling the H\'{e}non Map}

We implement the NG-RC-based controller by applying it to the H\'enon map. Here, the control perturbation is only applied to the $x$ variable, as $x$ is more sensitive to control than $y$.\cite{Braverman2022} Because we control a single variable, we adapt our notation introduced in Sec. \ref{sec:Control} by making ${\bf{Y}}\rightarrow Y$, ${\bf{u}}\rightarrow u$, ${\bf{W}}_u \rightarrow W_u$, ${\bf{K}} \rightarrow K$, and ${\bf{Y}_{des}}\rightarrow x_{des}$, as these quantities become scalars. 

As a first step toward the control algorithm implementation, we train the NG-RC to predict the dynamics of $x$ having current values for both $x$ and $y$, so that $ Y_{i+1}= x_{i+1}$ and ${\bf{X}}_i = [x_i,y_i]^T$. To choose the unique order-$p$ monomials, we follow the work of Wei et al. \cite{Wei_and_Billings_2004} who use system identification techniques on the H\'{e}non map to determine that order-2 monomials are sufficient to make accurate predictions. We use the unique second-order given by $\mathbb{O}_{nonlin,i} = [x_i^2,x_i y_i,y_i^2]^T$, and we choose $c=1$. The system identification approach assumes little to no information about the system and has also been used on more complex systems with higher dimensions to greatly reduce the number of nonlinear monomials.\cite{Wei_and_Billings_2004} Thus, it may be possible to use these techniques to construct computationally inexpensive NG-RC-based controllers for more complex systems, which we will explore in future work. Finally, after learning ${\bf{\hat W}}_{\bf{X}}$ and ${\hat W}_u$, the control signal is calculated according to the control law of Eq. \ref{eq:controlsignal}. 

Our focus lies on the control task of stabilizing the one-period orbit outside the trapping region, considering the presence of noise and modeling error. Additionally, we undertake the challenging task of stabilizing the four-period orbit, which is a known difficult task,\cite{Matousek2021} and controlling the system to an arbitrary trajectory.

\section{\label{sec:Results}Results}

We evaluate the effectiveness of our NG-RC-based controller using a variety of metrics. First, we evaluate the accuracy of the NG-RC prediction in the presence of noise. To quantify the prediction accuracy, we determine the root mean-square error (RMSE) between $x$ and $\hat x$ given by
 \begin{eqnarray}
 \text{RMSE}&=& \sqrt{{\langle{(x_i-{\hat x}_i)^2}\rangle}},
\label{eq:rmse}
\end{eqnarray}
where $\langle \cdot \rangle$ means averaging over many time steps, here taken to be $M_{test}$ steps.

\subsection{\label{sec: Prediction Robustness to Noise}Prediction Robustness to Noise}

To generate the training and testing data, an initial condition $(x_0,y_0)$ is chosen randomly with a uniform distribution inside the trapping region of the H\'{e}non Map. The map is iterated forward for a set number of steps, and a perturbation $g~u_i$ is applied at each step, $u_i$ is randomly sampled from a normal distribution with mean 0 and standard deviation 0.1, and $g$ is chosen to be 1 for simplicity. During the trajectory, the random perturbation can cause the system state to exit the trapping region and escape the attractor. In this case, a new initial condition is chosen and the current dataset is discarded. At the end of the trajectory, the dataset is split into a training and testing set, containing $M_{train}$ and $M_{test}$ data points, respectively. 

\begin{figure}[!ht]
\centering
	\subfloat{%
	\includegraphics[scale=0.47]{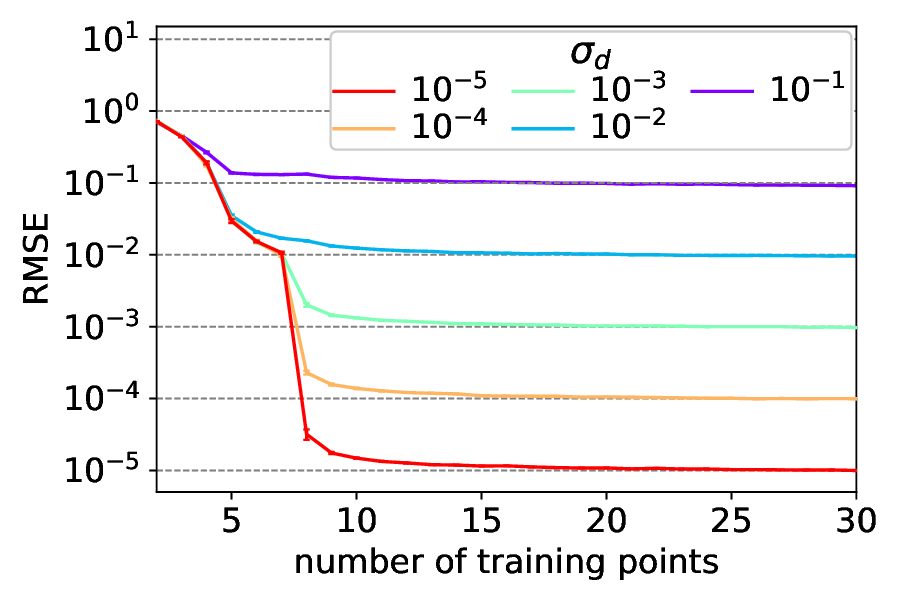}%
	}
 
\caption{Predicting the H\'{e}non map system for different noise levels (color coded) for limited training data. From top to bottom, the average RMSE after ten training points is $1.09 \pm 0.01 \times 10^{-5}$, $1.05 \pm 0.01 \times 10^{-4}$, $1.04 \pm 0.01 \times 10^{-3}$, $1.02 \pm 0.01 \times 10^{-2}$, and $0.98 \pm 0.01 \times 10^{-1}$, and the average optimal $\alpha$ is $9.93 \pm 0.02 \times 10^{-8}$, $8.11 \pm 0.02 \times 10^{-7}$, $6.89 \pm 0.02 \times 10^{-6}$, $1.62 \pm 0.65 \times 10^{-4}$, and $1.62 \pm 0.34 \times 10^{-2}$.
} 
\label{fig:sweep training points and noise rmse}
\end{figure}

For this first prediction task, we inject noise into the map through $d_{x,i}$ in Eq. \ref{eq:henon map xip1} and $d_{y,i}$ in Eq. \ref{eq:henon map yip1}. The noise is randomly sampled from a normal distribution with mean 0 and various standard deviations and $M_{train}$ is varied, while the regularization parameter is chosen $\alpha\in[0,1]$ to minimize the RMSE on $M_{test}=50$ (see the Appendix \ref{app:OptimizingAlpha}). The resulting curves are averaged over 100 initial conditions or trials (see Fig. \ref{fig:sweep training points and noise rmse}). The error bars in this figure, and throughout the paper, are derived from the standard deviation of the RMSE divided by the square root of the number of trials. The same metric is used when stating uncertainties for values of the RMSE and the regularization parameter. In Fig. \ref{fig:sweep training points and noise rmse}, the error bars are typically too small to be seen, indicating high confidence in the mean of the RMSE.

Looking at the low noise $\sigma_d = 10^{-5}$ case, the RMSE begins to quickly decrease after only two iterations, but reaches a critical point between iterations seven and eight where the RMSE begins to decrease much more rapidly. This is because the number of training data points is comparable to the number of learned model parameters and that the training data explore the extent of the attractor. The cases of higher noise levels initially display the same behavior, but gradually level off as the RMSE approaches the noise level. It is notable that the RMSE begins to approach to the noise level after $M_{train}=10$. This convergence demonstrates the NG-RC's ability to accurately predict the H\'{e}non map even in the presence of noise, while relying on a minimal number of data points. In all results that follow, the NG-RC is trained with $M_{train}=10$ without noise using random conditions for each trial, and the optimal $\alpha$ is zero.

\subsection{Controlling without Noise}

\begin{figure}[!ht]
\centering
	\subfloat{%
	\includegraphics[scale=0.33]{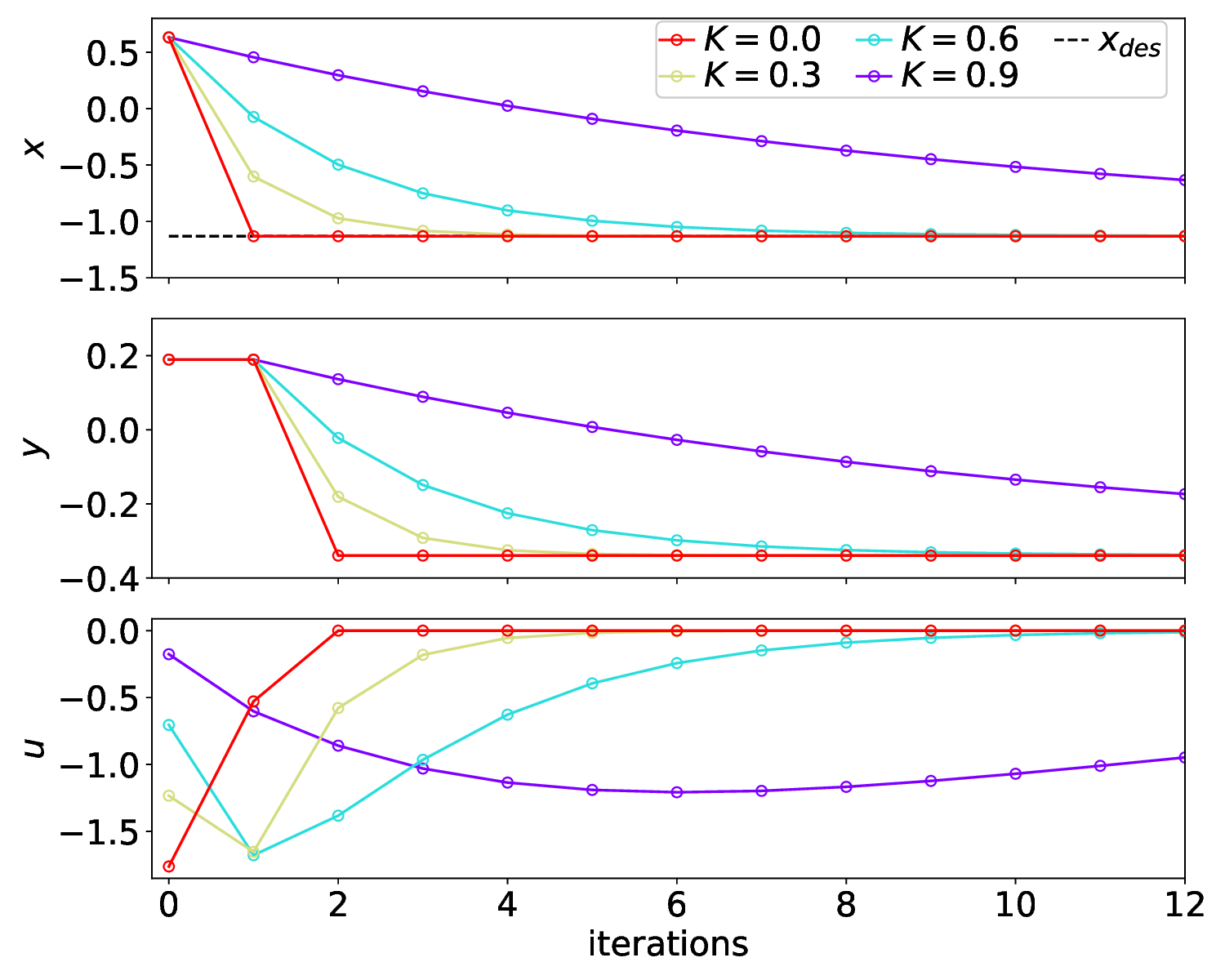}%
	}

\caption{Controlling the H\'{e}non map dynamics between the two unstable fixed points. (top) $x$, (middle) $y$, and (bottom) $u$ as a function of iterations for different values of $K$ (color code). The horizontal dashed line (top) indicates $x_{des}=x_{U2}$, and initial conditions for $x$ and $y$ are $x_{U1}$ and $y_{U1}$, respectively.} 

\label{fig:p1_orbit_control}
\end{figure}

We first investigate the task of controlling the H\'{e}non map from $P_{U1}$ to $x_{des} = x_{U2}$ to observe the effect of $K$ on the rate of convergence to the desired state. The result for this task without noise or modeling error can be seen in Fig. \ref{fig:p1_orbit_control}. For higher $K$ values, $x_i$ approaches $x_{des}$ slowly, while for lower $K$ values, it converges rapidly. In the $K = 0$ case, the relative error is $4.5 \times 10^{-15}$ after a single iteration. The low error can be attributed to Eq. \ref{eq:controlsignal} being driven solely by $x_{des,i+1}-\hat{x}_{i+1}$ when $K=0$, where $\hat{x}_{i+1}$ is a very good approximation of $x_{i+1}$ in the absence of perturbations. Furthermore, for $K=0.9$, $x_i$ approaches $x_{des}$ gradually with a relative error <1\% after 48 iterations (not shown in the plot). 

With the assumption of perfect learning and no disturbances, the tracking error is $e_{i+1} = 0.9 e_{i}$ according to Eq. \ref{eq:error_perfect}, meaning the error in the next step is 90\% of the current error. Given the initial error $x_{0} - x_{U2} = 1.7627$ in Fig. \ref{fig:p1_orbit_control}, a simple iteration of Eq. \ref{eq:error_perfect} also yields 48 iterations for a <1\% relative error, whose value agrees with the numerical simulation down to a precision of $6.8 \times 10^{-13}$. This agreement indicates that the assumption of perfect learning is valid, the control law is working as expected, and the NG-RC is a reliable model to learn this system. Note that even though we are only controlling $x_i$, $y_i$ is also driven toward $y_{U2}= 0.3x_{U2}$ because it is coupled to $x$.

\begin{figure}[!ht]
\centering
	\subfloat{%
	\includegraphics[scale=0.34]{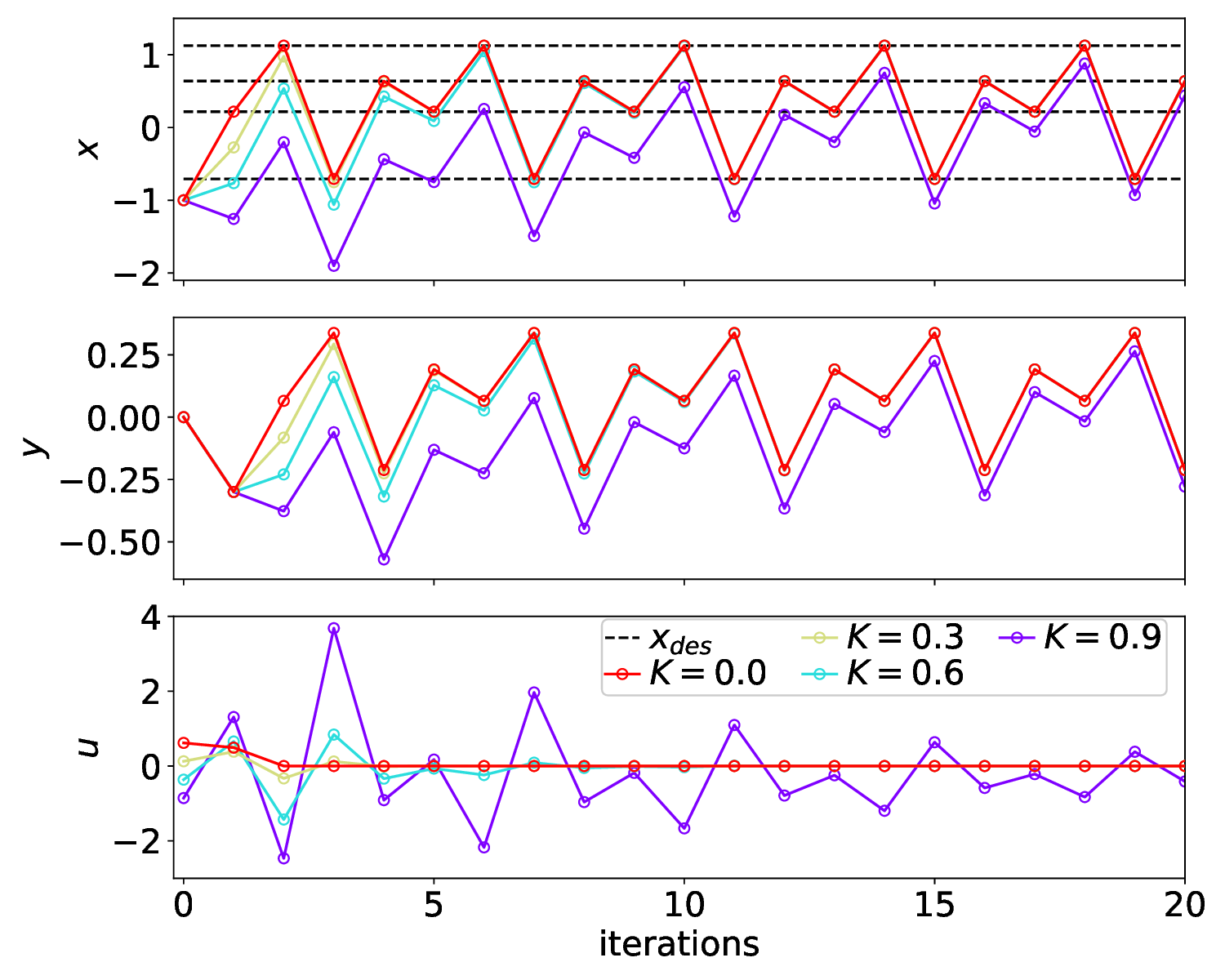}%
	}

\caption{Controlling H\'{e}non map dynamics to a four-period orbit. (top) $x$, (middle) $y$, and (bottom) $u$ as a function of iterations for different values of $K$ (color code). Horizontal dashed lines (top) indicate the $x_{des}$ values, which change between $P_1$, $P_2$, $P_3$, and $P_4$ every iteration step starting at $x_{des} = x_1$ at iteration 0.
} 
\label{fig:p4_orbit_control}
\end{figure}

A more difficult task is to control the system to the four-period orbit whose points are represented by the red circles in Fig. \ref{fig:attractor}. When the system is in the four-period orbit, it follows the trajectory $P_{1} \rightarrow P_{2} \rightarrow P_{3} \rightarrow P_{4} \rightarrow P_{1}$. As before, we seek to observe the effect of $K$ on the rate of convergence to the desired state. Because we only control $x$, we set $x_{des,i} = [x_1,x_2,x_3,x_4,x_1,...]$, and start at an initial condition $(x_0,y_0) = (-1,0)$, which is far from the points in the orbit. The result can be seen in Fig \ref{fig:p4_orbit_control}. As before, the larger $K$ values approach the desired values gradually, while the smaller $K$ values approach the desired values rapidly. The $K = 0.0$ case achieves a relative error of $1.8 \times 10^{-12}$ by the first iteration, and the $K = 0.9$ case achieves a relative error <1\% by iteration 50 (not shown). As before, $y$ approaches those of the four-period orbit, even though $y$ is not controlled explicitly. We conclude that even for the difficult task of controlling the system to a four-period orbit, the control law performs as expected. 

\begin{figure}[!ht]
\centering
	\subfloat{%
	\includegraphics[scale=0.34]{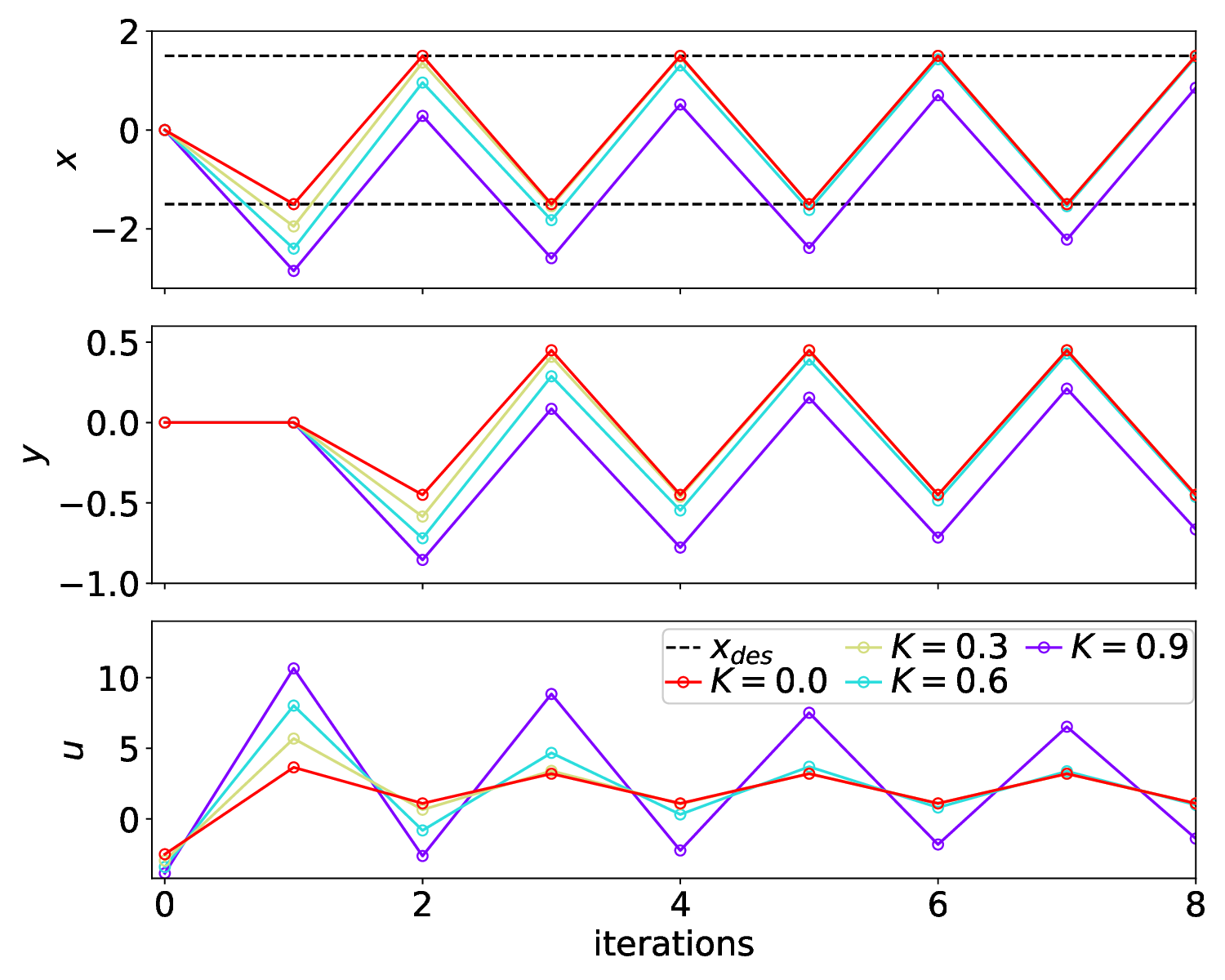}%
	}

\caption{Controlling H\'{e}non map dynamics between two arbitrary values $x=-1.5$ and $x=1.5$. The first desired value is $x_{des} = -1.5$. (top) $x$, (middle) $y$, and (bottom) $u$ as a function of iterations for different values of $K$ (color code). Horizontal dashed lines (top) indicate the $x_{des}$ values. The initial condition is set to $(x_0,y_0) =(0,0)$.
} 
\label{fig:arbitrary_control}
\end{figure}

To show that the control law is not limited to unstable periodic orbits, we evaluate the performance on an arbitrary control task, controlling the system repeatedly between $x=-1.5$, and $x=1.5$. Both of these points are outside of the trapping region, meaning that large control perturbations are required to keep the system from diverging to infinity. The result can be seen in Fig. \ref{fig:arbitrary_control}. As with the other control tasks, the $K=0$ trajectory reaches the desired value in one iteration with a relative error equal to $2.88 \times 10^{-5}$, and the $K=0.9$ trajectory reaches the desired value with a relative error <1\% after iteration 44. Thus, we conclude that the NG-RC-based controller can control the system to an arbitrary desired state, which is not possible for many controllers.

\subsection{Control in the Presence of Noise and Modeling Error}

To show that the NG-RC-based controller is robust to unexpected perturbations, we inject noise into the H\'{e}non map during control. The noise is injected in both $x$ and $y$ variables using $d_{x}$ and $d_{y}$ as shown in Eqs. \ref{eq:henon map xip1}-\ref{eq:henon map yip1}. The noise is sampled from a normal distribution with zero mean. We vary the noise strength in the same manner as in Sec. \ref{sec: Prediction Robustness to Noise}. 

\begin{figure}[!ht]
\centering
	\subfloat{%
	\includegraphics[scale=0.5]{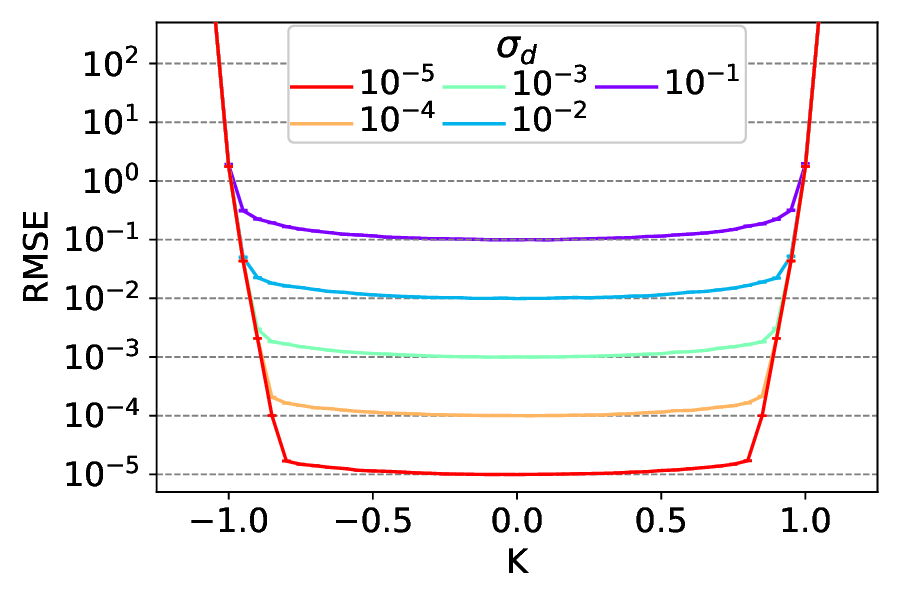}%
	}

\caption{Mean RMSE between $x$ and $x_{des}$ for the control task $P_{U1}\rightarrow P_{U2}$ as a function of $K$ for different noise levels (color code). From top to bottom, the average minimum RMSE is $0.99 \pm 0.01 \times 10^{-5}$, $1.00 \pm 0.01 \times 10^{-4}$, $1.00 \pm 0.01 \times 10^{-3}$, $0.99 \pm 0.01 \times 10^{-2}$, and $0.99 \pm 0.01 \times 10^{-1}$.
}
\label{fig:sweep_K_and_noise_rmse}
\end{figure}

To determine the effectiveness of the NG-RC-based controller subject to noise, we repeat the task of controlling the system between the two unstable fixed points $P_{U1} \rightarrow P_{U2}$, but we choose to focus on the stability of the system after it has reached $P_{U2}$ and compare the error to the noise level. To do this, we vary $K$ and calculate the RMSE between iterations 50 and 150 for each and average over 100 trials. If the RMSE for a given trial and K is greater than $10^3$, the system is diverging to infinity, and we consider it a failure and exclude it from the figures. For this task, there are no failures for $|K| \leq 0.9$. As shown in Fig. \ref{fig:sweep_K_and_noise_rmse}, the RMSE rapidly increases when $|K|$ approaches 1, indicating that the controller is unstable for $|K|>1$ for all noise levels in agreement with Sarangapani's stability analysis.\cite{Sarangapani2006} Note that the RMSE approaches the noise level when $K \rightarrow 0$, which is what we expect from Eq. \ref{eq:error2} in the case of perfect learning, as the tracking error becomes the noise. The reason the error does not remain a constant for $|K|<1$ and then diverge to infinity at $|K|=1$ is because 150 iterations are not enough for the error to converge to the noise level for all $|K|<1$. For example, we can iterate the tracking error in Eq. \ref{eq:error_perfect} with $K = 0.99$ using initial error $x_{0} - x_{U2} = 1.7627$ to find the error should reach the noise level of $\sigma_{d} = 10^{-5}$ in 1202 iterations. Similar to Fig. \ref{fig:sweep training points and noise rmse}, the error bars are typically too small to be seen, indicating a high confidence in the mean of the RMSE. 

\begin{figure}[!ht]
\centering
	\subfloat{%
	\includegraphics[scale=0.5]{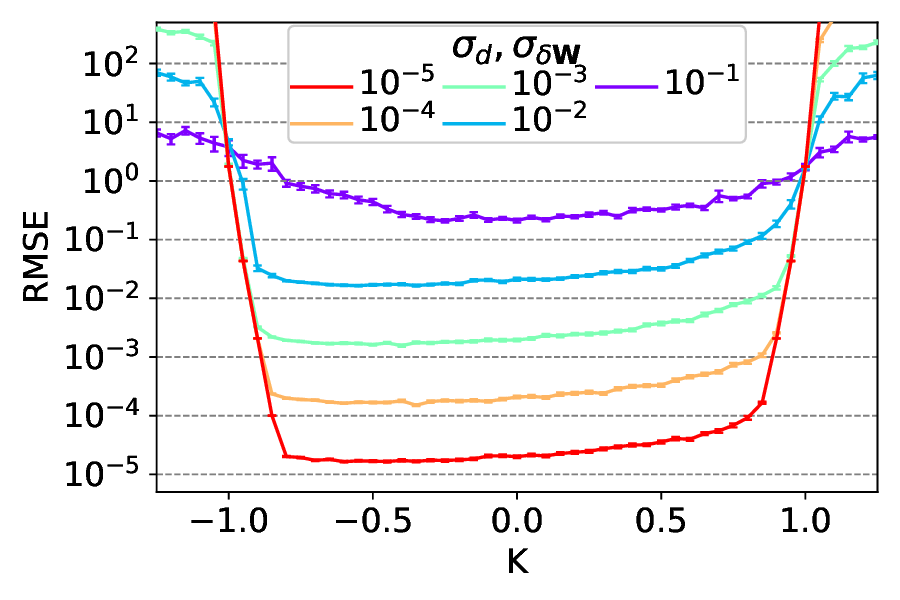}%
	}

\caption{Mean RMSE between $x$ and $x_{des}$ for the control task $P_{U1}\rightarrow P_{U2}$ as a function of $K$ for different noise levels and modeling error (color code). From top to bottom, the average minimum RMSE is $1.64 \pm 0.05 \times 10^{-5}$, $1.51 \pm 0.06 \times 10^{-4}$, $1.56 \pm 0.07 \times 10^{-3}$, $1.64 \pm 0.05 \times 10^{-2}$, and $2.08 \pm 0.14 \times 10^{-1}$.
} 
\label{fig:sweep_K_and_noise_and_modeling_rmse}
\end{figure}

Although the modeling error of the NG-RC is very small for learning the H\'{e}non map, this might not be true for every system. To explore this possibility, we include modeling error to our controller and perform the same $P_{U1}\rightarrow P_{U2}$ control task as in Fig. \ref{fig:sweep_K_and_noise_rmse}. The modeling error is added to the NG-RC by perturbing the learned weights after the training phase with the additive term ${\bf{\delta W}}$ so that ${\bf{\hat W}} \rightarrow {\bf{\hat W}} + {\bf{\delta W}}$. This perturbation is equivalent to the additive modeling error $\epsilon$ shown in Eq. \ref{eq:MLoutputFxWu}. Note that ${\bf{\delta W}}$ has as many components as ${\bf{\hat W}}$, each of which is randomly sampled from a normal distribution with zero mean and standard deviation $\sigma_{\bf{\delta W}}$ and held constant during the control phase. Here, we choose $\sigma_{\bf{\delta W}} = \sigma_{d}$, i.e., both the noise and the modeling perturbation have the same strength. As shown in Fig. \ref{fig:sweep_K_and_noise_and_modeling_rmse}, which is obtained using the same procedures used to obtain Fig. \ref{fig:sweep_K_and_noise_rmse}, the controller displays robustness to both noise and modeling error similar to when only noise is present (Fig. \ref{fig:sweep_K_and_noise_rmse}). Here, the RMSE dependence on $K$ is less smooth than in the previous case. One possible reason is that the control law is strongly based on the learned weights, which implies that the controller noise sensitivity increases with the modeling error. The minimum RMSE for this task is roughly $1.6$ times larger in comparison with the case where only noise is applied, which is expected because the controller has two sources of noise. Moreover, the range of $K$ where the control is stable is preserved. 

Using the same metric for failure as the previous control task, the control experiences failures for 0.16\% of trials for $\sigma_{d},\sigma_{\bf{\delta W}} = 10^{-2}$, and $|K| \leq 0.9$, 11.1\% of trials for $\sigma_{d},\sigma_{\bf{\delta W}} = 10^{-1}$, and $|K| \leq 0.9$, while there are no failures for the other parameters. These results indicate that the NG-RC-based controller is robust to random disturbances.


\section{\label {sec:Discussions}Discussion}

In this work, we demonstrate the effectiveness of the NG-RC for controlling the H\'{e}non map when using it in the formalism developed by Sarangapani.\cite{Sarangapani2006} Our results demonstrate several key advantages when compared to other ML-based control methods. 

We highlight that the NG-RC-based controller requires significantly fewer training points when compared to other ML approaches. The NG-RC is able to learn the dynamics of the H\'{e}non map with as few as ten data points even with injected noise, while other previously reported methods require 100$\times$ more training data to perform similar control tasks.\cite{Gokce2013,Garcia2022} Additionally, the RMSE during prediction is approximately equal to the noise level, which, for the $\sigma_{d} = 10^{-5}$ case, is 1000$\times$ smaller than the RMSE seen in other studies without noise.\cite{Leite2021}

We demonstrate that the NG-RC-based controller excels at three challenging control tasks. High gain allows the system to reach the desired state in a single iteration, while low gain allows for a tunable rate of convergence. This is in contrast to other methods that require over ten iterations to stabilize to the desired state for similar tasks.\cite{Senkerik2011,Leite2021} Matousek \textit{et al.}\cite{Matousek2021} show similar convergence times for the four-period orbit task using a technique known as genetic programming, but their best solution uses delayed points and polynomials in the feature vector up to the sixth order, which we do not require. 

We demonstrate that the NG-RC-based controller is robust to noise and modeling errors, showing that it can control the system to a desired state with an error close to the noise level. Real-world systems are often subject to unpredictable disturbances, and our model showcased its ability to handle such challenges. These qualities are necessary for reliability and performance in practical applications. 

An important aspect of our approach is computational efficiency. Our model requires only seven weights to successfully control the H\'{e}non map, allowing the control law to be computed at a low computational cost. Furthermore, optimizing the control is an analytical procedure that uses regularized regression (see the Appendix \ref{app:OptimizingAlpha}). In contrast, other control methods, such as evolutionary algorithms, require iterative optimization to determine the weights, which can be computationally expensive. Some models require expensive optimization before each control action, as in model predictive control. The NG-RC has the same predictive power but is paired with a much simpler approach. This indicates that the NG-RC is suitable for hardware implementation and edge computing, where real-time control is often required. 

The NG-RC has very few hyperparameters to optimize compared to other ML methods, and appropriate parameter values can often be obtained based on physical insights. We demonstrate this by incorporating the dynamics of the H\'{e}non map into the design of the polynomial features to achieve an optimal control law. Previous studies have also employed other physical principles, such as symmetry considerations, to achieve optimal predictions.\cite{Barbosa2021} Thus, information about the underlying dynamics of the target system has the potential to accelerate nonlinear controller design using NG-RC. If little information about the system is available, system identification techniques can be used to achieve the optimal polynomial features, which has been shown to work for the H\'{e}non map and higher-dimensional systems with more complex dynamics.\cite{Wei_and_Billings_2004}

While our work shows the NG-RC-based controller can stabilize discrete-time maps, future work will explore the application of our controller to discretely sampled continuous systems, which will allow further comparisons with other reservoir computer based controllers.\cite{Canaday2021,Haluszczynski2021,haluszczynski_2023} The NG-RC has already been proven to excel at forecasting such systems, \cite{NGRC,NGRC-L96} and Sarangapani's work is designed specifically for such systems. Furthermore, we seek to test the performance of the NG-RC-based controller on real-world systems by implementing the controller in hardware, where we can take advantage of the simplicity of the control law and the parallelizability of the polynomial features to control systems on short time scales. Additionally, it has been shown that Tikhonov regularization can be performed iteratively,\cite{Hertz1991} allowing for the possibility of an NG-RC-based controller that can continuously adapt to changing conditions, which reflects realistic control situations. This could also allow for more direct comparisons with reservoir computing based controllers that stabilize systems subject to external factors or order parameter changes.\cite{Haluszczynski2021,haluszczynski_2023}

In conclusion, the NG-RC-based controller excels at controlling the H\'{e}non map in a variety of challenging tasks, learning the dynamics of the system with small training dataset sizes and is robust to noise and modeling error. The previous success of the NG-RC in modeling dynamical systems makes this a promising approach for controlling higher-dimensional systems with more complex dynamics.

\section*{Acknowledgement}
This material is based upon work supported by the Air Force Office of Scientific Research under award number FA9550-22-1-0203.

\section*{Data Availability Statement}

The data that support the findings of this study are available from the corresponding author upon reasonable request.

\appendix

\section{\label{app:OptimizingAlpha}Optimizing the NG-RC for Prediction}

To compute $\bf{\hat{W}}$, we use Tikhonov regularization,\cite{Vogel_2002} which has the explicit form
\begin{align}
    \bf{\hat{W}} = \bf{Y}_{d}\mathbb{O}_{total}^T (\mathbb{O}_{total} \mathbb{O}_{total}^T + \alpha \bf{I})^{-1},
\end{align}
where $\bf{Y}_{d}$ contains $\bf{Y}_{i+1}$, and $\mathbb{O}_{total}$ contains $\mathbb{O}_{total,i}$
for all $i$ in $M_{train}$ training points, and $\alpha$ is the regularization parameter, also known as ridge parameter, used to prevent over-fitting. Once the optimal $\hat{{\bf{W}}}$ is obtained in training phase, prediction is made using Eq. \ref{eq:MLoutput} for a given initial state.

\bibliography{aipsamp}

\end{document}